\documentclass[11pt]{article}

\usepackage[final]{acl}

\usepackage{times}
\usepackage{latexsym}

\usepackage[T1]{fontenc}
\usepackage[utf8]{inputenc}

\usepackage{microtype}

\usepackage{inconsolata}

\usepackage{graphicx}

\usepackage{enumitem}
\usepackage{amssymb}
\usepackage{amsmath}
\usepackage{xcolor}
\usepackage[table]{xcolor}
\usepackage{pifont} % 用于显示勾选和叉号符号
\usepackage{booktabs}
\usepackage{multirow}
\title{ICT-NLP at SemEval-2026 Task 3: Less Is More — Multilingual Encoder with Joint Training and Adaptive Ensemble for Dimensional Aspect Sentiment Regression}

\author{
 \textbf{Liyuan Huang\textsuperscript{1,2,3}},
 \textbf{Jiawei He\textsuperscript{1,2}},
 \textbf{Wutao Shen\textsuperscript{1,2,3}},
 \textbf{Lin Li\textsuperscript{1,2}},
 \textbf{Jin Zhang\textsuperscript{1,2}}
\\
 \textsuperscript{1}State Key Laboratory of AI Safety\\
 \textsuperscript{2}Institute of Computing Technology, Chinese Academy of Sciences\\
 \textsuperscript{3}University of Chinese Academy of Sciences
\\
\texttt{\{huangliyuan25e,shenwutao25e,lilin2020,jinzhang\}@ict.ac.cn} \\
\texttt{hepisces@qq.com}
}

\begin{document}

\maketitle

\begin{abstract}
This paper describes our system to SemEval-2026 Task 3 Track A Subtask 1 on Dimensional Aspect Sentiment Regression (DimASR). We propose a lightweight and resource-efficient system built entirely on multilingual pre-trained encoders, without relying on LLMs or external corpora. 
We adopt joint multilingual and multi-domain training to facilitate cross-lingual transfer and alleviate data sparsity, introduce a bounded regression transformation that improves training stability while constraining predictions within the valid range, and employ an adaptive ensemble strategy via subset search to reduce prediction variance.
Experimental results demonstrate that our system achieves strong and consistent performance, ranking 1st on \textit{zho-res}, 2nd on \textit{zho-lap}, and 3rd on \textit{jpn-hot}, with all remaining datasets placed within the top half of participating teams\footnotemark.

\footnotetext{Our code is available at \url{https://github.com/huangly312/SemEval2026-task3-DimASR}.}
\end{abstract}

\section{Introduction}
\label{sec:introduction}

Aspect-based sentiment analysis (ABSA) aims to identify the sentiment expressed toward specific aspects in text \cite{pontiki-etal-2014-semeval,zhang-etal-2022-survey}.
Most existing ABSA studies formulate the task as coarse-grained classification (e.g., \textit{positive}, \textit{negative}, \textit{neutral}), which cannot capture fine-grained affective distinctions such as those between \textit{good} and \textit{excellent}.
Drawing on the theory from affective science \cite{russell1980circumplex}, sentiment can be represented along fine-grained, real-valued dimensions of \textbf{valence} (negative–positive) and \textbf{arousal} (sluggish–excited).

\begin{figure}[t]
    \centering
    \includegraphics[width=\linewidth]{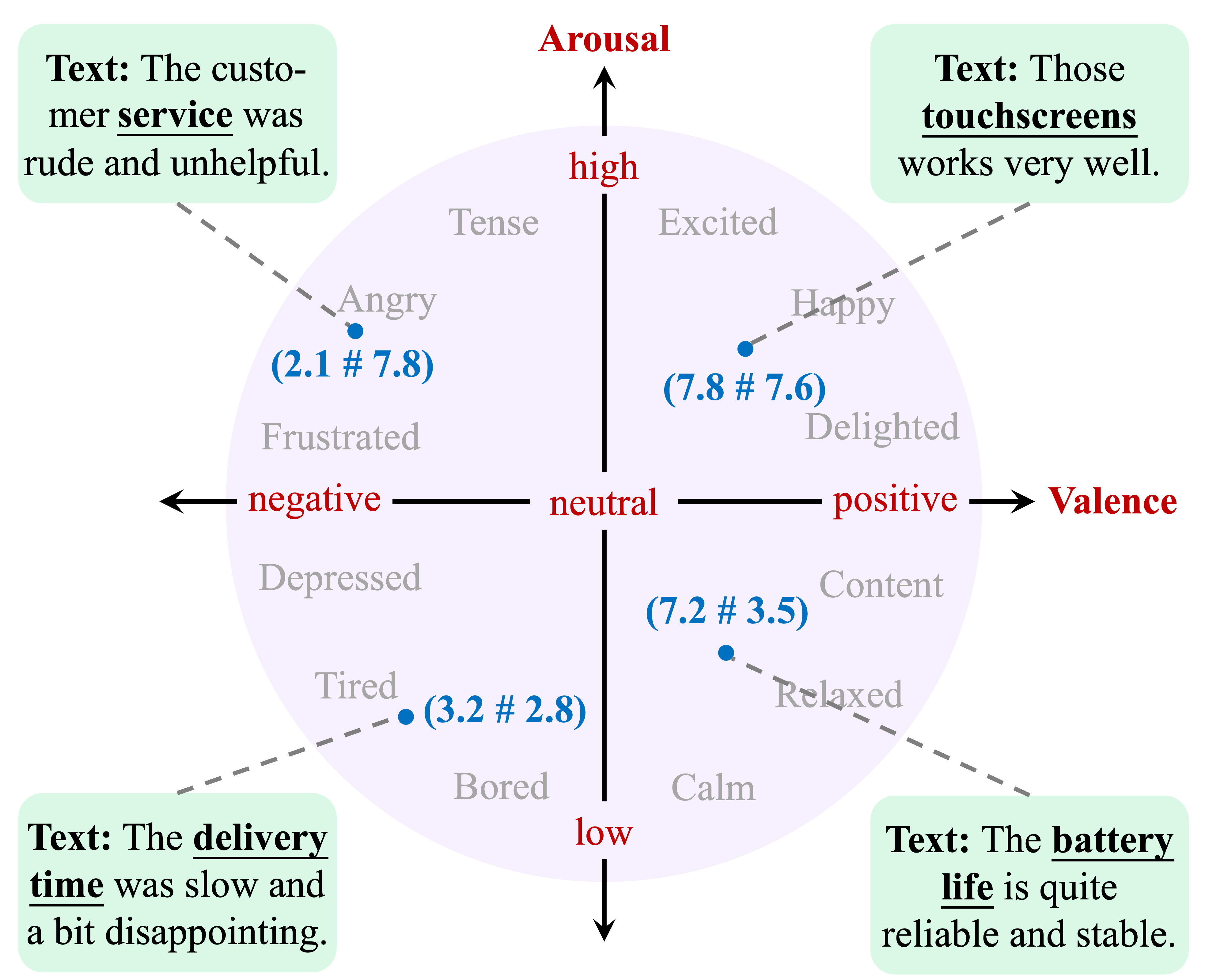}
    \caption{Illustration of Subtask~1 (DimASR).}
    \label{fig:task}
\end{figure}

The DimABSA shared task \cite{yu-etal-2026-semeval} introduces a multilingual and multi-domain benchmark 
that integrates dimensional sentiment analysis into the traditional ABSA framework.
We focus on Track A Subtask~1, Dimensional Aspect Sentiment Regression (DimASR), which requires predicting continuous valence--arousal (VA) scores (1--9 scale) for a given aspect within a text (Figure~\ref{fig:task}).

While large language models (LLMs) have become a dominant strategy for sentiment tasks, such approaches often involve high computational cost and limited reproducibility. 
In contrast, we investigate how competitive a lightweight system can be without using LLMs.

In this paper, we present a lightweight and resource-efficient system built entirely on multilingual pre-trained encoders without leveraging LLMs, external corpora, or data augmentation techniques. 
We adopt a joint multilingual and multi-domain training strategy to facilitate cross-lingual transfer and alleviate data scarcity. 
We further introduce a sigmoid-based bounded output transformation to improve training stability and ensure predictions within the valid range. 
Finally, we apply an adaptive ensemble strategy via exhaustive search to select the optimal subset from the candidate model pool for each specific language--domain pair, significantly reducing prediction variance.

Our system achieves strong performance across all 10 language--domain datasets, ranking 1st on \textit{zho-res}, 2nd on \textit{zho-lap}, and 3rd on \textit{jpn-hot}, with all remaining datasets placed within the top half of participating teams. 
These results suggest that, in dimensional sentiment regression, less is more: a carefully designed lightweight encoder-based system without LLMs can remain highly competitive against approaches that demand far greater computational resources.

\section{Related Work}
\label{sec:Related Work}

Prior work on dimensional sentiment analysis has focused on building affective resources and developing regression models across multiple granularities.
On the resource side, sentiment lexicons provide word-level VA norms \cite{warriner-etal-2013-norms,mohammad-2018-obtaining}, while sentence-level corpora offer broader contextual coverage \cite{preotiuc-pietro-etal-2016-modelling,buechel-hahn-2017-emobank}. Chinese dimensional resources have also been developed to address the gap in non-English coverage \cite{yu-etal-2016-building,lee2022chinese}. On the modeling side, early approaches relied on LSTM-based architectures for VA prediction \cite{wu-etal-2017-thu,cheng2021valence,wang-etal-2016-dimensional}, while pre-trained Transformers subsequently became the dominant paradigm. For instance, by employing pre-trained BERT and MLP for V-A prediction,~\cite{xu-etal-2024-hitsz}, or incorporating contrastive learning approaches~\cite{tong-wei-2024-cciiplab}.
More recently, LLM-based methods have shown strong performance on dimensional sentiment tasks, either via in-context learning(ICL) or supervised fine-tuning(SFT) \cite{xu-etal-2024-hitsz}, and represent the prevailing approach in recent shared task competitions.

\section{System Overview}
\begin{figure}[ht]
    \centering
    \includegraphics[width=\linewidth]{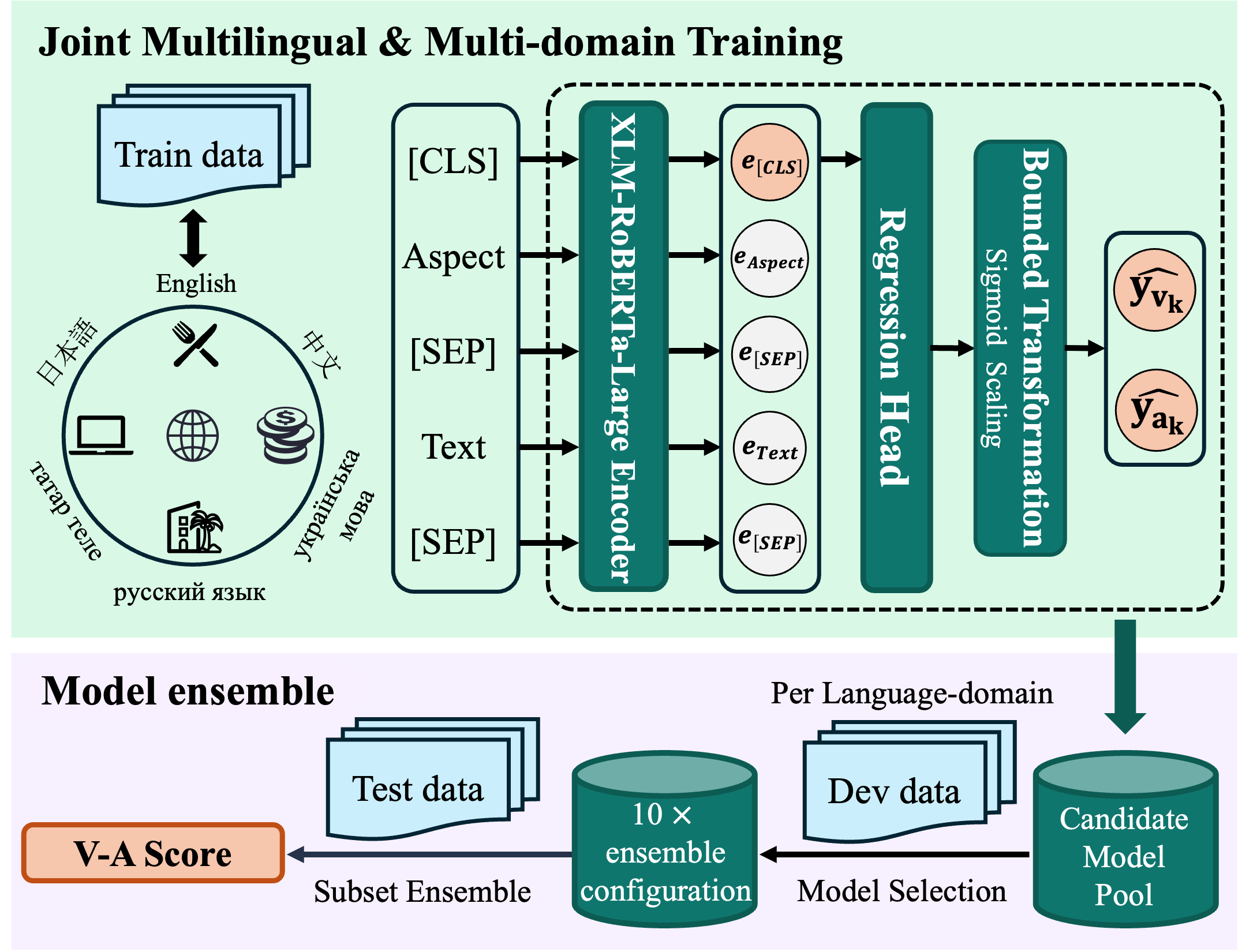}
    \caption{The architecture of our DimASR system.}
    \label{fig:architecture}
\end{figure}

Figure~\ref{fig:architecture} illustrates the overall pipeline of our system. Given a text and its associated aspect, we encode them as a sentence pair using a multilingual pre-trained encoder, and predict the VA scores via a regression head. Models are trained jointly across all language--domain pairs, and the final predictions are obtained via a development-set-guided adaptive ensemble.

%----------------------------------------------
\subsection{Data Processing}

The training data is provided in a quadruplet format (\textit{Aspect}, \textit{Category}, \textit{Opinion}, \textit{VA}) shared across all subtasks of Track~A. Since we focus on Subtask~1 (DimASR), we extract only the (\textit{Text}, \textit{Aspect}, \textit{VA}) fields and apply several preprocessing steps before training.

Following the official dataset note that the test set excludes implicit \texttt{NULL} annotations, we remove all training instances where the aspect is \texttt{NULL} to ensure consistency between training and test distributions.
We also discard 3 instances with VA values outside the valid range $[1, 9]$.
For instances with multiple aspect terms, we expand them into independent samples, each paired with the full review text and a single aspect term along with its corresponding VA score.
For instances where multiple opinions are associated with the same aspect, we retain only the VA score corresponding to the first opinion.
%----------------------------------------------

\subsection{Model Architecture}
\label{sec:Model Architecture}

\paragraph{Backbone Encoders.}
We explore three multilingual pre-trained encoders as the backbone: mBERT~\citep{devlin-etal-2019-bert}, XLM-RoBERTa-base, and XLM-RoBERTa-large~\citep{conneau-etal-2020-unsupervised}. These models provide strong cross-lingual representations and are well suited to our multilingual setting. In addition, these models span a range of scales and architectures, allowing us to examine the effect of model size on cross-lingual sentiment regression.

\paragraph{Input Representation.}
The aspect term and review text are encoded as a sentence pair. Depending on the backbone, the tokenizer inserts model-specific special tokens: mBERT formats the input as \texttt{[CLS] aspect [SEP] text [SEP]}, while XLM-RoBERTa uses \texttt{<s> aspect </s></s> text </s>}. All sequences are truncated or padded to a maximum length of 128 tokens.

\paragraph{Regression Head.}
We take the hidden state of the first special token (\texttt{[CLS]} or \texttt{<s>}) as the sentence-level representation.
A dropout layer is applied before a linear projection that maps the hidden representation to two raw output values corresponding to valence and arousal:
\vspace{-0.15cm}
\begin{equation}
    \mathbf{h} = \mathrm{Dropout}(\mathbf{e}_{\texttt{[CLS]}}), \quad
    \hat{\mathbf{y}} = \mathbf{W}\mathbf{h} + \mathbf{b},
    \label{eq:head}
\end{equation}
where $\mathbf{W} \in \mathbb{R}^{2 \times d}$ and
$\mathbf{b} \in \mathbb{R}^{2}$ are learnable parameters,
and $d$ is the hidden size of the backbone.
The model is trained by minimizing the Mean Squared Error (MSE) loss, which is consistent with the official evaluation metric $\text{RMSE}_{\mathrm{VA}}$ (Equation \ref{eq:rmse_va}).

\paragraph{Bounded Output Transformation.}
The official task defines a valid VA range of $[1, 9]$. 
To align the model's output space with this requirement and enhance training stability, we further apply a sigmoid-based bounded transformation to the raw regression output:
\vspace{-0.1cm}
\begin{equation}
    \hat{\mathbf{y}}_{\mathrm{bounded}}
        = \sigma\!\left(\hat{\mathbf{y}}\right) \times 8 + 1,
    \label{eq:sigmoid}
\end{equation}
where $\sigma(\cdot)$ denotes the sigmoid function.

This ensures that all predictions are valid at inference time and facilitates smoother convergence during the early stages of training.
During our experiments, we observe that although the transformation stabilizes the training process and yields a consistent overall improvement, its effect varies across individual settings. Therefore, we treat it as a flexible component and include both bounded and unbounded model variants in the candidate pool for the subsequent ensemble stage to leverage their complementary strengths.

% -----------------------------------------------------------
\subsection{Joint Multilingual and Multi-Domain Training}
We adopt a joint multilingual and multi-domain training strategy, pooling data from all language--domain pairs to train a single unified model. Each instance is represented solely by its text--aspect pair, without explicit language or domain indicators. 

Compared with language–domain specific training, which learns independent models for each pair, joint training mitigates overfitting in low-resource settings and encourages the model to learn more generalizable representations that facilitate cross-lingual and cross-domain knowledge transfer \cite{conneau-etal-2020-unsupervised,VanThin03042023}. In particular, for language–domain pairs with limited training instances, the shared encoder is exposed to a substantially larger and more diverse set of sentiment expressions, allowing low-resource pairs to benefit from representations learned from richer languages and domains, thereby compensating for the lack of in-domain supervision.

% -----------------------------------------------------------
\subsection{Model Ensemble}
\label{sec:ensemble}

To reduce prediction variance and improve robustness, we adopt a structured ensemble strategy.

\paragraph{Candidate Pool Construction.}
We evaluate all trained models on the development sets across language--domain pairs and select 7 checkpoints that consistently achieve strong overall performance while collectively covering the top-performing models for each individual pair, forming the candidate pool for ensemble.
All 7 candidates are built on the XLM-RoBERTa-large backbone, but differ in batch size, learning rate, number of training epochs, and whether the sigmoid output constraint (Equation~\ref{eq:sigmoid}) is applied, providing ensemble diversity beyond hyperparameter variation alone. Detailed configurations of the candidate models are provided in Appendix~\ref{app:ensemble_model}.

\paragraph{Ensemble Subset Selection.}
Rather than a naive uniform averaging, we perform an exhaustive search over all possible subsets of size 2 to 7 drawn from the candidate model pool for each language--domain pair independently.
For each subset, predictions are obtained by averaging the valence and arousal outputs of the selected models element-wise.
The subset that achieves the lowest RMSE$_{\mathrm{VA}}$ on the development set is selected as the final ensemble configuration for that specific pair. 
As the optimal subset varies across pairs, this process results in 10 distinct ensemble configurations, which are subsequently applied to the official test sets for final submission. 
The selected optimal subsets for each language--domain pair are reported in Appendix~\ref{app:ensemble_config_pair}.

This exhaustive yet controlled ensemble selection strategy effectively reduces prediction variance and consistently improves performance across all language--domain pairs.

\section{Experimental Setup}
\label{sec:setup}
\paragraph{Dataset.}
The official DimABSA Track~A dataset \cite{lee2026dimabsabuildingmultilingualmultidomain} covers 6 languages (Chinese, English, Japanese, Russian, Tatar, and Ukrainian) and 4 domains (Hotel, Laptop, Restaurant, and Finance), organized into 10 language--domain pairs for Subtask~1 (DimASR).

In our experiments, we utilize the official training and development splits, and reserve 10\% of the training data as an internal validation set. 

%----------------------------------------------
\paragraph{Evaluation Metrics.}
According to the official task guidelines, Subtask~1 (DimASR) is evaluated by measuring the prediction error in the VA space using Root Mean Squared Error (RMSE). The metric is defined as:
\vspace{-0.2cm}
\begin{equation}
\small
\mathrm{RMSE}_\mathrm{{VA}} =
\sqrt{
\frac{1}{N}
\sum_{i=1}^{N}
\left(V_{p}^{(i)}-V_{g}^{(i)}\right)^2+
\left(A_{p}^{(i)}-A_{g}^{(i)}\right)^2
}
\label{eq:rmse_va}
\end{equation}
where $N$ is the number of instances; $V_p^{}$ and $A_p^{}$ denote the predicted valence and arousal values for an instance; and $V_g^{}$ and $A_g^{}$ denote the corresponding gold values.

%----------------------------------------------
\paragraph{Implementation Details.}
Our models are implemented using PyTorch 2.3 and Hugging Face Transformers 4.46.1.
All experiments are conducted on a single NVIDIA A800 GPU (40GB VRAM).
The model parameters are optimized using AdamW with early stopping based on validation RMSE, where the patience is set to 2. 
To promote model diversity for ensemble, we train multiple models under different hyperparameter configurations. 
Specifically, we vary the batch size in $\{16, 32, 64\}$ and the learning rate within the range of $8\times10^{-6}$ to $3\times10^{-5}$.
All experiments are conducted with a fixed random seed of 42.

%----------------------------------------------
\paragraph{Baselines.}
We compare our method with the official baseline results reported by the task organizers \cite{lee2026dimabsabuildingmultilingualmultidomain}. 
These baselines include LLM-based approaches evaluated in zero-shot and few-shot settings, such as \textbf{Kimi K2 Thinking} \cite{team2025kimi}, as well as supervised fine-tuned LLMs, including \textbf{Qwen-3 14B} \cite{yang2025qwen3} and \textbf{GPT-OSS 120B} \cite{agarwal2025gpt}. 

\section{Results}
\subsection{Main Results}
\begin{table*}[ht]
\centering
\footnotesize
\setlength{\tabcolsep}{3.5pt}
\renewcommand{\arraystretch}{1.1}
\begin{tabular}{l cccccccccc >{\columncolor[gray]{0.95}}c}
\toprule
Methods 
& eng-res & eng-lap & jpn-hot & jpn-fin 
& rus-res & tat-res & ukr-res 
& zho-res & zho-lap & zho-fin 
& \textbf{Avg.} \\
\midrule

\multicolumn{12}{l}{\textit{\textbf{Zero-Shot Learning}}} \\
Kimi K2 Thinking$^\dagger$ 
& 2.3432 & 2.6546 & 2.3294 & 2.3379 
& 2.0630 & 2.3636 & 2.0782 
& 2.6230 & 2.0426 & 2.9662 
& 2.3802 \\

\midrule
\multicolumn{12}{l}{\textit{\textbf{One-Shot Learning}}} \\
Kimi K2 Thinking$^\dagger$ 
& 2.1461 & 2.1893 & 1.7553 & 1.6396 
& 1.7768 & 1.9380 & 1.7805 
& 1.8959 & 1.6440 & 1.9652 
& 1.8731 \\

\midrule
\multicolumn{12}{l}{\textit{\textbf{Supervised Fine-Tuning}}} \\
Qwen-3 14B$^\dagger$       
& 2.6427 & 2.8089 & 2.2906 & 1.8964 
& 2.1528 & 2.6367 & 2.2121 
& 2.0073 & 1.7706 & 1.4707 
& 2.1889 \\

GPT-OSS 120B$^\dagger$     
& \underline{1.4605} & \underline{1.5269} & \underline{0.7188} & \underline{1.0188} 
& \underline{1.4775} & \textbf{1.7153} & \underline{1.5166} 
& \underline{1.0349} & \underline{0.8032} & \underline{0.6511} 
& \underline{1.1924} \\

\midrule
\rowcolor[gray]{0.95}
\textbf{Ours}    
& \textbf{1.2676} & \textbf{1.3098} & \textbf{0.6289} & \textbf{0.8331} 
& \textbf{1.4420} & \underline{1.8596} & \textbf{1.4750} 
& \textbf{0.9256} & \textbf{0.6553} & \textbf{0.4892} 
& \textbf{1.0886} \\

\bottomrule
\end{tabular}
\caption{
Track~A Subtask~1 Results. $\text{RMSE}_{\mathrm{VA}}$ on the official test sets across 10 language--domain pairs.
The \textbf{best} results are in bold, and the \underline{second-best} are underlined.
Baseline results$^\dagger$ from~\cite{lee2026dimabsabuildingmultilingualmultidomain}.
}
\label{tab:main_result}
\end{table*}
Table~\ref{tab:main_result} reports the final performance of our system on the official test sets across all 10 language--domain pairs, in comparison with the official baselines.
Our system outperforms the strongest supervised fine-tuning baseline, GPT-OSS 120B, on 9 out of 10 language–domain pairs.
Moreover, it consistently surpasses both the one-shot prompted closed-source LLM Kimi K2 Thinking and the QLoRA fine-tuned Qwen3-14B across all pairs.

On the official leaderboard, our system ranks 1st on \textit{zho-res}, 2nd on \textit{zho-lap}, and 3rd on \textit{jpn-hot}, with all remaining datasets placed within the top half of participating teams. These results demonstrate that a lightweight PLM-based system, when equipped with joint multilingual training and adaptive ensemble selection, can be highly competitive against LLM-based approaches without relying on external corpora or data augmentation.

% ---------------------------------

\subsection{Ablation Study}
\begin{table*}[ht]
\centering
\footnotesize
\setlength{\tabcolsep}{3.5pt} % 略微增加间距
\renewcommand{\arraystretch}{1.1} % 增加行高，更美观
\begin{tabular}{l cccccccccc >{\columncolor[gray]{0.95}}c} % 为最后一列添加浅灰色背景突出平均值
\toprule
Variant & eng-res & eng-lap & jpn-hot & jpn-fin & rus-res & tat-res & ukr-res & zho-res & zho-lap & zho-fin & \textbf{Avg.} \\
\midrule
\multicolumn{12}{l}{\textit{\textbf{1. Training Strategy (mBERT)}}} \\
Separate Training   & 1.3086 & 1.2863 & 1.1860 & 1.2455 & 1.8048 & 2.1028 & 1.8087 & 0.8985 & 0.8467 & 0.6697 & 1.3158 \\
Joint Training  & \textbf{1.1537} & \textbf{1.2715} & \textbf{1.1590} & \textbf{1.2090} & \textbf{1.5532} & \textbf{1.6845} & \textbf{1.4242} & \textbf{0.8048} & \textbf{0.8223} & \textbf{0.6482} & \textbf{1.1730} \\
\rowcolor[gray]{0.95} 
\textit{Relative Change} 
& \textcolor{blue}{-11.8\%} 
& \textcolor{blue}{-1.2\%} 
& \textcolor{blue}{-2.3\%} 
& \textcolor{blue}{-2.9\%} 
& \textcolor{blue}{-13.9\%} 
& \textcolor{blue}{-19.9\%} 
& \textcolor{blue}{-21.3\%} 
& \textcolor{blue}{-10.4\%} 
& \textcolor{blue}{-2.9\%} 
& \textcolor{blue}{-3.2\%} 
& \textbf{\textcolor{blue}{-10.9\%}} \\
\midrule
\multicolumn{12}{l}{\textit{\textbf{2. Backbone Model Size (Joint Training)}}} \\
mBERT               & 1.1537 & 1.2715 & 1.1590 & 1.2090 & 1.5532 & 1.6845 & 1.4242 & 0.8048 & 0.8223 & 0.6482 & 1.1730 \\
XLM-R Base          & 1.1498 & 1.1072 & 0.9672 & 0.9663 & 1.5050 & 1.8109 & 1.4582 & 0.7790 & 0.7951 & 0.6048 & 1.1144 \\
XLM-R Large         & \textbf{1.0892} & \textbf{1.0383} & \textbf{0.9416} & \textbf{0.9286} & \textbf{1.4000} & \textbf{1.6519} & \textbf{1.4198} & \textbf{0.6987} & \textbf{0.7550} & \textbf{0.5808} & \textbf{1.0504} \\
\midrule
\multicolumn{12}{l}{\textit{\textbf{3. Bounded Output (Joint Training; XLM-R Large)}}} \\
w/o sigmoid & 1.0892 & 1.0383 & 0.9416 & 0.9286 & 1.4000 & 1.6519 & 1.4198 & 0.6987 & 0.7550 & 0.5808 & 1.0504 \\
w/ sigmoid  & 1.1294 & 0.9978 & 0.9157 & 0.8299 & 1.3598 & 1.6599 & 1.3258 & 0.7420 & 0.6763 & 0.4838 & 1.0120 \\
\rowcolor[gray]{0.95} \textit{Relative Change} & \textcolor{red}{+3.7\%} & \textcolor{blue}{-3.9\%} & \textcolor{blue}{-2.8\%} & \textcolor{blue}{-10.6\%} & \textcolor{blue}{-2.9\%} & \textcolor{red}{+0.5\%} & \textcolor{blue}{-6.6\%} & \textcolor{red}{+6.2\%} & \textcolor{blue}{-10.4\%} & \textcolor{blue}{-16.7\%} & \textbf{\textcolor{blue}{-3.7\%}} \\
\midrule
\multicolumn{12}{l}{\textit{\textbf{4. Inference Strategy}}} \\
Best Single Model   & 0.9299 & 0.9287 & 0.8823 & 0.7898 & 1.2777 & 1.5109 & 1.3198 & 0.6614 & 0.6763 & 0.4838 & 0.9461 \\
Ensemble   & \textbf{0.9165} & \textbf{0.9020} & \textbf{0.8568} & \textbf{0.7601} & \textbf{1.2648} & \textbf{1.4397} & \textbf{1.2661} & \textbf{0.6492} & \textbf{0.6509} & \textbf{0.4766} & \textbf{0.9183} \\
\rowcolor[gray]{0.95} \textit{Relative Change} & \textcolor{blue}{-1.4\%} & \textcolor{blue}{-2.9\%} & \textcolor{blue}{-2.9\%} & \textcolor{blue}{-3.8\%} & \textcolor{blue}{-1.0\%} & \textcolor{blue}{-4.7\%} & \textcolor{blue}{-4.1\%} & \textcolor{blue}{-1.8\%} & \textcolor{blue}{-3.8\%} & \textcolor{blue}{-1.5\%} & \textbf{\textcolor{blue}{-2.9\%}} \\
\bottomrule
\end{tabular}
\caption{Ablation study on the official development set (RMSE$_{\text{VA}}$). Bold font indicates the best performance within each comparison block.}
\label{tab:ablation}
\end{table*}
To validate the contribution of each component in our framework, we conduct a series of ablation experiments on the official development set. Results are reported in Table~\ref{tab:ablation}.

\paragraph{Joint Multilingual Training.}
Taking mBERT as a representative backbone, we observe that joint training significantly outperforms separate training across all pairs, yielding an overall average RMSE reduction of 10.9\%. 
Notably, the gains are most pronounced for low-resource language-domain pairs, such as \textit{ukr-res} (-21.3\%) and \textit{tat-res} (-19.9\%), whereas improvements are comparatively modest for higher-resource pairs such as \textit{eng-lap} (-1.2\%) and \textit{zho-lap} (-2.9\%). This asymmetric pattern suggests that pooling data across languages and domains provides implicit supervision for low-resource pairs, effectively alleviating data sparsity through shared cross-lingual representations.

\paragraph{Backbone Model Size.}
Block~2 compares three backbone encoders under joint training. Scaling from mBERT to XLM-R Large results in a consistent performance boost, demonstrating that the superior cross-lingual representation capability of XLM-R Large provides a more robust foundation for capturing the subtle nuances of valence and arousal dimensions.

\paragraph{Bounded Output Transformation.}
We examine the effect of the sigmoid-based bounded transformation (Equation \ref{eq:sigmoid}) by using a representative configuration (XLM-R Large, batch size 16, learning rate 1e-5), and observe an average RMSE reduction of 3.7\%. Although the gains are not uniform across all language--domain pairs, the bounded transformation generally stabilizes training and prevents out-of-range predictions. In the final ensemble, we include both bounded and unbounded models to balance robustness and flexibility, leveraging their complementary strengths.

\paragraph{Ensemble Strategy.}
Block 4 shows that the adaptive subset ensemble consistently outperforms the best single model across all 10 datasets (avg.\ $-$2.9\%), confirming that exhaustive subset search over the candidate pool effectively reduces prediction variance.

\section{Conclusion}
In this paper, we present a lightweight and resource-efficient system for Subtask 1 (Dimensional Aspect Sentiment Regression) of SemEval-2026 Task 3 Track~A. Without relying on LLMs, external corpora, or data augmentation, our approach leverages joint multilingual training and an adaptive ensemble strategy.

Extensive experiments across 10 language--domain pairs demonstrate that, in dimensional sentiment regression, less is more: a carefully designed PLM-based system remains highly competitive against strong LLM-based baselines.

\section{Ethical Considerations}
All data used in this work are sourced from the official competition release and are used solely for scientific research purposes in strict accordance with the provided data usage agreements. Our system does not involve sensitive personal information. AI-assisted tools were used only for language polishing, while all research ideas, experimental design, and conclusions were independently developed and verified by the authors.

% 显示所有参考文献
% \nocite{*}
\bibliography{ref}

\clearpage
\onecolumn
\appendix
\section*{Appendix}
\section{Candidate Model Hyperparameter Configurations}
\label{app:ensemble_model}

\begin{table}[ht]
\centering
\small
\renewcommand{\arraystretch}{1.1} % 稍微加高行间距，更显精致
\setlength{\tabcolsep}{3pt}
\begin{tabular}{lcccc}
\toprule
\textbf{Model ID} & \textbf{Batch Size} & \textbf{Learning Rate} & \textbf{Epoch} & \textbf{Sigmoid} \\
\midrule
\rowcolor[gray]{0.95}
M1 & 16 & 1e-5 & 7 & \checkmark \\
M2 & 32 & 1e-5 & 3 &            \\
\rowcolor[gray]{0.95}
M3 & 32 & 1e-5 & 5 & \checkmark \\
M4 & 32 & 1e-5 & 7 & \checkmark \\
\rowcolor[gray]{0.95}
M5 & 32 & 2e-5 & 5 & \checkmark \\
M6 & 32 & 8e-6 & 3 & \checkmark \\
\rowcolor[gray]{0.95}
M7 & 32 & 8e-6 & 7 &            \\
\bottomrule
\end{tabular}
\caption{Hyperparameter configurations of the seven candidate models. All models utilize XLM-RoBERTa-large as the backbone and are trained under a joint multilingual and multi-domain setting.}
\label{tab:ensemble_models}
\end{table}

% ----------------------------------------
\section{Ensemble Selection Results}
\label{app:ensemble_config_pair}

\begin{table}[h]
\centering
\small
\renewcommand{\arraystretch}{1.1} % 稍微加高行间距，更显精致
\setlength{\tabcolsep}{3.5pt}    % 调整列间距
% \begin{tabular}{lccccccc >{\columncolor[gray]{0.95}}c} % 为最后一列加底色突出
\begin{tabular}{lcccccccc}
\toprule
\textbf{Pair} & \textbf{M1} & \textbf{M2} & \textbf{M3} & \textbf{M4} & \textbf{M5} & \textbf{M6} & \textbf{M7} & \textbf{Number} \\
\midrule
\rowcolor[gray]{0.95} % 斑马纹底色
eng-lap  &           &           &           & \checkmark & \checkmark &           &           & 2 \\
eng-res  &           &           &           & \checkmark &           & \checkmark &           & 2 \\
\rowcolor[gray]{0.95}
jpn-fin  & \checkmark &           & \checkmark & \checkmark &           &           & \checkmark & 4 \\
jpn-hot  &           &           & \checkmark &           &           & \checkmark & \checkmark & 3 \\
\rowcolor[gray]{0.95}
rus-res  & \checkmark &           &           &           & \checkmark &           &           & 2 \\
tat-res  &           & \checkmark &           & \checkmark &           &           & \checkmark & 3 \\
\rowcolor[gray]{0.95}
ukr-res  & \checkmark &           &           &           & \checkmark &           &           & 2 \\
zho-fin  & \checkmark &           &           & \checkmark &           &           & \checkmark & 3 \\
\rowcolor[gray]{0.95}
zho-lap  & \checkmark &           &           &           & \checkmark & \checkmark & \checkmark & 4 \\
zho-res  &           & \checkmark & \checkmark &           &           &           & \checkmark & 3 \\
\bottomrule
\end{tabular}
\caption{Selected ensemble configurations for each language--domain pair. \checkmark~indicates that the corresponding candidate model is included in the final ensemble.}
\label{tab:ensemble_config_pair}
\end{table}

\end{document}